\pdfoutput=1
\documentclass[letterpaper,10pt,conference]{ieeeconf}
\IEEEoverridecommandlockouts
\usepackage[hyphens]{url}
\usepackage{graphicx}
\usepackage{amsmath,amssymb,amsfonts}
\usepackage{dsfont}
\usepackage{mathtools}
\usepackage{bm}
\usepackage{booktabs}
\usepackage{array}
\usepackage{multirow}
\usepackage{makecell}
\makeatletter
\let\labelindent\relax
\makeatother
\usepackage{enumitem}
\usepackage[table]{xcolor}
\usepackage{pifont}
\usepackage{microtype}
\usepackage{algorithm}
\usepackage{algpseudocode}
\usepackage[hidelinks]{hyperref}
\usepackage{cleveref}
\usepackage{flafter}
\usepackage{placeins}
\usepackage{balance}
\usepackage{dblfloatfix}

\newcommand{\citep}[1]{\cite{#1}}

\graphicspath{{figures/}}
\hypersetup{
  pdfauthor={Wanli Liuchen, Fangyuan Wang, Bin Li, Anqing Duan, Yunhui Liu, Peng Zhou, David Navarro-Alarcon},
  pdftitle={Phase-and-First-Arrival VLM Feedback for Sparse-Reward Reinforcement Learning in Surgical Manipulation}
}
\begin{document}

% =================================================================
\title{\LARGE\bf
Phase-and-First-Arrival VLM Feedback for\\Sparse-Reward Reinforcement Learning in Surgical Manipulation}

\author{{\small Wanli Liuchen$^{1}$, Fangyuan Wang$^{1}$, Bin Li$^{2}$, Anqing Duan$^{3}$, Yunhui Liu$^{2}$, Peng Zhou$^{4}$, and David Navarro-Alarcon$^{1}$}\\[-1pt]
{\scriptsize $^{1}$The Hong Kong Polytechnic University; $^{2}$The Chinese University of Hong Kong;}\\[-2pt]
{\scriptsize $^{3}$Mohamed bin Zayed University of Artificial Intelligence; $^{4}$Great Bay University;}\\[-2pt]
{\scriptsize Corresponding author: David Navarro-Alarcon (dnavar@polyu.edu.hk).}}

\maketitle
\vspace{-8pt}
\begingroup
\centering
\scriptsize\itshape
This work has been submitted to IEEE for possible publication. Copyright may transfer without notice, after which this version may no longer be accessible.\par
\endgroup
\vspace{5pt}
\thispagestyle{empty}
\pagestyle{empty}

% =================================================================
\begin{abstract}
% =================================================================
Sparse outcome feedback limits what robots can learn from unsuccessful attempts at complex manipulation. Failed multi-stage surgical attempts can contain grasps, lifts, or transfers worth reusing. In sparse-reward reinforcement learning, terminal rewards collapse such attempts to the same outcome, while scalar vision--language model (VLM) ratings reveal neither what progress merits credit nor when it occurred. We introduce phase-and-first-arrival feedback: one VLM query per recorded episode identifies the furthest visually verified task phase and when that phase is first reached, allowing the learner to reuse partial behavior and localize credit. We instantiate it in SurgPhaseBench, a phase-structured suite spanning rigid and deformable tasks, and evaluate it in simulation and hardware. Across five simulated tasks, our method reaches $75.2\%$ mean success, compared with $52.1\%$ for a reward based on Contrastive Language--Image Pre-training (CLIP) using the same visual input; the advantage persists when only the feedback representation changes. On hardware, the same record supports autonomous block picking and slip recovery. Together, these results show that trajectory-level visual supervision can preserve partial progress while providing the temporal credit needed for sparse-reward control.
\end{abstract}

\noindent\textbf{Project website---}\url{https://surgphase.verloge.space}

\noindent\textbf{Index Terms---}Surgical Robotics: Laparoscopy, Reinforcement Learning, Computer Vision for Medical Robotics.

% =================================================================
\section{Introduction}
\label{sec:intro}
% =================================================================

% Single-column teaser: declared at the start of the Introduction so it
% lands at the top of the page-1 right column; placement verified in the
% RA-L build.
\begin{figure}[t]
  \centering
  \includegraphics[width=\columnwidth]{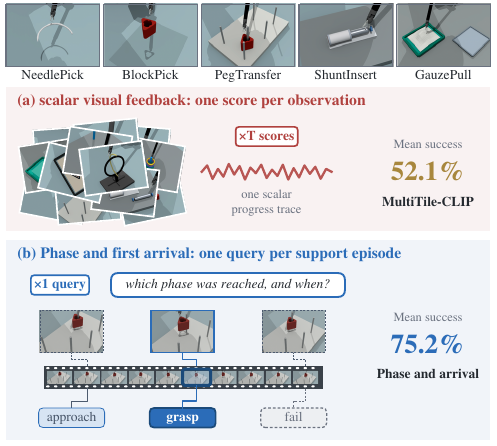}
  \caption{One structured query bridges episode- and transition-level supervision. A confidence-and-progress gate retains partial attempts, rank scales their credit, and first arrival localizes reuse. After 120k policy updates, phase-conditioned replay reaches 75.2\% mean success versus 52.1\% for a CLIP scalar reward using the same views.}
  \label{fig:teaser}
\end{figure}

Surgical autonomy requires reliable policies for precise, contact-rich manipulation across ordered stages~\citep{bendikas2023needle,ho2025surgirl}. Reinforcement learning offers one route, but a terminal success reward gives every unsuccessful attempt the same return~\citep{thananjeyan2020saved,bendikas2023needle}. A needle-picking attempt that grasps and lifts the needle before dropping it earns no more credit than one that never makes contact, yet contains behavior worth reusing. Dense shaping recovers that signal, but it is authored per task and reads privileged simulator state that hardware does not expose, so the effort does not carry to a new procedure or to a real manipulator. The challenge is to preserve partial progress without task-specific dense rewards.

Vision--language models supply feedback that needs no task-specific predicate: they judge progress from images and a short task description~\citep{rocamonde2024zeroshot,du2023success,wang2024rlvlmf,luu2025erlvlm}. The forms in use, however, trade economy against resolution. A scalar trajectory rating costs one query per episode but reports only how good the attempt was, not which event earned that score or when it happened. Pairwise preferences rank two attempts without locating progress inside either. Framewise scoring does localize, but only among the observations it scores, so finer temporal resolution is bought with proportionally more queries. Reinforcement learning needs the opposite of a summary: credit attached to particular transitions. Can one query per episode retain the economy of trajectory-level feedback while still locating reusable behavior in time?

We address this with phase-and-first-arrival feedback over ordered surgical phases. Peg Transfer progresses through grasps, handoffs, and placements~\citep{fried2004proving}, and surgical workflow models label procedural video by phase~\citep{twinanda2016endonet,rueckert2025phakir}. In our tasks, these phases are visually distinguishable, so they can be verified without state predicates; their order gives even an unsuccessful episode measurable partial progress. In \Cref{fig:teaser}, the VLM reads a completed-episode storyboard and returns the furthest verified phase and the interval in which it is first reached: phase identifies reusable progress, and arrival locates it in time.

A fixed compiler turns each record into a learning signal. A confidence-and-progress gate rejects uncertain or no-progress records; phase rank scales retained credit; and first arrival localizes it to an event-centered window for replay, critic targets, and behavior cloning. The VLM processes a fixed pool of recorded support episodes once before policy learning, producing a frozen record bank for mixing with online experience. Collection and evaluation continue on the sparse environment reward.

We instantiate the method in SurgPhaseBench, twelve phase-structured ManiSkill3 environments spanning rigid and deformable interaction: five evaluate the complete observer--compiler--learner loop, and seven establish executable and semantic breadth. Across the five policy-learning tasks, phase-and-first-arrival feedback outperforms matched scalar feedback. The same frozen-record interface supports autonomous block picking and recovery from slips on physical hardware, with no VLM query during policy deployment.

Our contributions are:
\begin{enumerate}[leftmargin=1.5em,itemsep=1pt,topsep=2pt]
  \item We introduce phase-and-first-arrival feedback, a VLM-based method that converts one storyboard query per support episode into a structured progress-and-timing record, then compiles that record into localized credit for reusing partial behavior from unsuccessful attempts.
  \item We provide SurgPhaseBench, a twelve-environment benchmark for phase-structured, sparse-reward surgical manipulation across rigid and deformable interactions, with a five-task track for end-to-end policy learning.
  \item We show that phase-and-first-arrival feedback improves five-task mean success by $7.4$ percentage points over scalar feedback from the same VLM. On a physical robot, the learned block-picking policy succeeds in 18 of 20 held-out trials without deployment-time VLM queries.
\end{enumerate}

% =================================================================
\section{Related Work}
\label{sec:related}
% =================================================================

\paragraph{Surgical platforms and benchmarks}
Open surgical learning platforms include dVRL, SurRoL, LapGym, and Orbit-Surgical~\citep{richter2019open,xu2021surrol,scheikl2023lapgym,yu2024orbit}. They provide subtasks and support demonstration-constrained RL, sim-to-real manipulation, autonomous suction, and multi-task automation~\citep{thananjeyan2020saved,scheikl2023sim2real,ou2024suction,ho2025surgirl}. SurgPhaseBench addresses a complementary evaluation need: each environment couples sparse success with ordered visual milestones and first-arrival semantics, making partial progress a common observable across tasks.

\paragraph{Phases and task structure}
Bendikas et al.~\citep{bendikas2023needle} define exploratory bottlenecks and train nested NeedlePickAndPlace subtasks. Their decomposition changes the training problem into a sequence of subtasks. Our pipeline preserves the original episode and stores its furthest phase and first-arrival interval as replay metadata. Workflow recognition predicts framewise phase sequences for video understanding~\citep{twinanda2016endonet,rueckert2025phakir}. Our observer instead returns one maximal phase and arrival interval for replay and policy learning.

\paragraph{Visual supervision for control}
Visual models can score observations or trajectory videos~\citep{rocamonde2024zeroshot,baumli2024rewards,sontakke2023roboclip}. SuccessVQA locates the first frame after which success persists and clones accepted behavior~\citep{du2023success}; RL-VLM-F and ERL-VLM collect online preferences or ratings, while PLARE learns from fixed VLM preferences without a reward model~\citep{wang2024rlvlmf,luu2025plare,luu2025erlvlm}. These produce success, scalar, or preference supervision. VICtoR, REDS, and SARM instead learn dense or stage-aware rewards from demonstrations or labels~\citep{hung2024victor,kim2025reds,chen2025sarm}. Our pipeline uses one query on each fixed support episode to jointly identify partial progress and its first arrival, then compiles the records before policy learning.

\paragraph{Structured credit and replay}
Reward machines require event propositions evaluable along a trajectory~\citep{icarte2018rewardmachines}. HER relabels transitions with state-accessible achieved goals~\citep{andrychowicz2017her}; SurRoL's HER+DEMO applies Q-filtered cloning to demonstrations~\citep{xu2021surrol,nair2018overcoming}. Potential shaping redistributes a specified reward~\citep{ng1999potential}. Our post-hoc visual record instead turns one verified partial phase and its arrival into a transition distribution, localized critic targets, and Q-filtered reuse while retaining the original task goal and sparse environment reward.

% =================================================================
\section{Methodology}
\label{sec:method}
% =================================================================

\begingroup
\setlength{\abovedisplayskip}{2.25pt plus 1pt minus 1pt}
\setlength{\belowdisplayskip}{2.25pt plus 1pt minus 1pt}
\setlength{\abovedisplayshortskip}{2pt plus 1pt}
\setlength{\belowdisplayshortskip}{3pt plus 1pt minus 1pt}
\setlength{\jot}{2pt}

\begin{figure*}[!t]
  \centering
  % (FEEDBACK_INTERFACE=1); no PDF masking or text overlay is used.
  % The two lettered insets are Sec. III-A and Sec. III-B; the loop they sit in
  % is the ground, so the caption has to name it -- neither carries a title.
  \includegraphics[width=\textwidth]{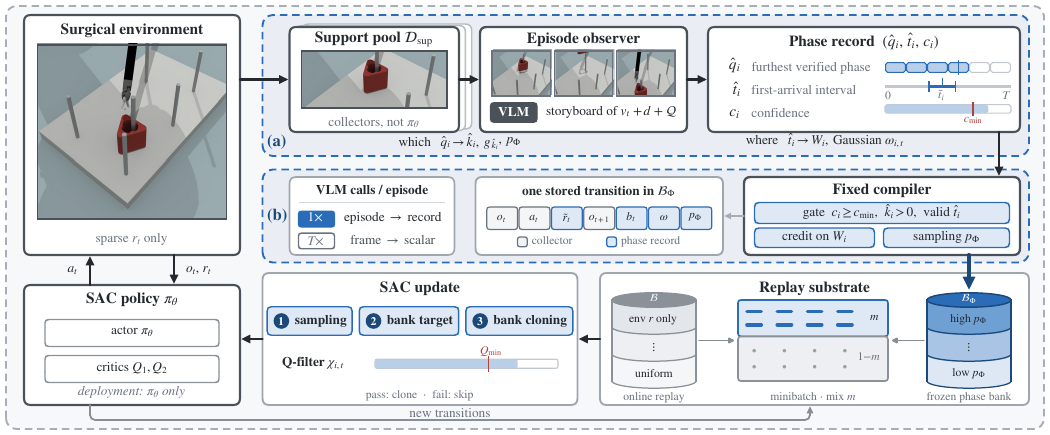}
  \caption{From a completed attempt to reusable replay. The surrounding loop collects and evaluates on the sparse environment reward, while compiled support episodes form a frozen bank. (a)~One query per support episode returns a phase record. (b)~A fixed compiler produces bank rewards and sampling/BC weights; the online learner applies the Q-filter. The bank is mixed into each minibatch at rate $m$.}
  \label{fig:system-overview}
\end{figure*}

\subsection{Problem Formulation}
\label{sec:formulation}

We consider finite-horizon continuous-control episodes with sparse environment reward $r_t$ and terminal bit $b_t$. A Soft Actor-Critic (SAC) policy $\pi_\theta(a\mid o)$~\citep{haarnoja2018sac} acts from state observation $o_t$, and synchronized RGB $v_t$ provides the post-hoc visual record. We target settings where unsuccessful episodes share one return, so grasping and lifting before a drop is indistinguishable from no contact.

The input is a support pool frozen before learning, with each transition sequence paired to synchronized RGB. From episode-level feedback, we recover \emph{which} unsuccessful episodes contain reusable behavior and \emph{where} it occurs. \Cref{fig:system-overview} separates the resulting stages: one frozen-VLM query and deterministic compilation per support episode, then mixed-replay SAC. Online collection and evaluation use $r_t$; only bank samples receive auxiliary credit.

\subsection{Episode-Level Phase Observation}
\label{sec:observer}

For task $\mathcal T$, the author supplies description $d_{\mathcal T}$, a cue-annotated ordered vocabulary $\mathcal Q_{\mathcal T}=(q_0\prec\cdots\prec q_{K_{\mathcal T}})$, and a nondecreasing map $\rho_{\mathcal T}:\mathcal Q_{\mathcal T}\to\{0,\ldots,K\}$ with $\rho_{\mathcal T}(q_0)=0$. Task-specific labels map to shared progress ranks; authors specify their observable semantics and order.

The five simulated task orders are BlockPick/NeedlePick: \emph{clear failure, approach, contact or grasp, lift, stable success}; PegTransfer: \emph{clear failure, approach, contact or grasp, lift, transfer, stable success}; ShuntInsert: \emph{insert failure, approach, grasp, align, seat}; and GauzePull: \emph{clear failure, approach, attach, pull, target align, target reach}. Their semantics map monotonically to the shared ranks \emph{none, approach, grasp, lift, at-goal, success}; inapplicable ranks are skipped.

The fixed support pool $\mathcal D_{\mathrm{sup}}=\{\tau_i\}_{i=1}^{N}$ contains episodes $\tau_i$ of length $T_i$, each pairing a transition sequence with synchronized RGB, and may include failed attempts with reusable partial behavior. A fixed adapter $A_{\mathcal T}$ turns the episode video into a timestamped storyboard with full views and task-specified crops. The phase vocabulary emphasizes visually verifiable milestone states, such as an object being lifted, transferred, aligned, or placed. By combining chronological context with full views and task-relevant crops, the storyboard lets the observer identify the furthest state evidenced at any time in the episode, while adjacent timestamps delimit a first-arrival bracket sufficient for localized credit assignment. When evidence supports two adjacent labels, the observer conservatively selects the lower phase. The frozen observer is queried once:
\[
  (\hat q_i,\hat t_i,c_i)=
  f_\phi\!\left(A_{\mathcal T}(\tau_i),d_{\mathcal T},\mathcal Q_{\mathcal T}\right),
  \qquad \hat k_i=\rho_{\mathcal T}(\hat q_i),
\]
The observer reads the storyboard and task specification. Here $\hat q_i$ is the furthest verified phase, $\hat t_i\in\{[\hat t_{i,\ell},\hat t_{i,u}],\varnothing\}$ its first-arrival bracket, and $c_i\in[0,1]$ its self-reported confidence for gating. If arrival falls between displayed times, the observer returns their step interval; it returns $\varnothing$ when timing is not identifiable, and $q_0$ always carries $\varnothing$. Maximal phase identifies partial progress; first arrival excludes repeats.

\subsection{Compiling Phase Records into Replay}
\label{sec:hindsight-integration}

\begin{figure}[!t]
  \centering
  % Base: phase_anatomy_framework_v8.pdf; current notation overlay:
  \includegraphics[width=\columnwidth]{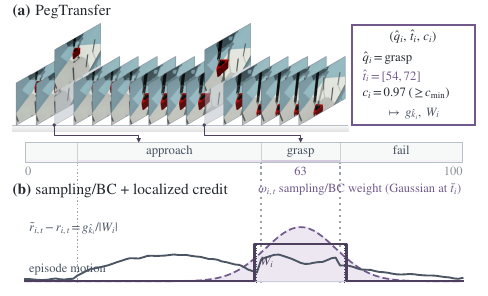}
  \caption{Phase compilation on one recorded attempt. (a)~The observer maps a completed simulated peg-transfer episode to furthest phase $\hat q$, first-arrival interval $\hat t$, and confidence $c$; $\rho_{\mathcal T}$ maps the label to shared rank $\hat k$. (b)~The gate retains verified progress, rank sets total bank credit, and the arrival midpoint defines a credit window and Gaussian sampling/BC weights (\Cref{eq:compiled-credit,eq:arrival-weight}).}
  \label{fig:phase-anatomy}
\end{figure}

As illustrated in \Cref{fig:phase-anatomy}, the gate accepts a record if $c_i\ge c_{\min}$, $\hat k_i>0$, and $0\le\hat t_{i,\ell}\le\hat t_{i,u}<T_i$. The compiler uses midpoint $\bar t_i=(\hat t_{i,\ell}+\hat t_{i,u})/2$ and clipped window $W_i=\{t:0\le t<T_i,\,|t-\bar t_i|\le h\}$, where $h\ge\tfrac12$ ensures $W_i$ is nonempty. Let $0=g_0\le\cdots\le g_K$ be the shared total-credit schedule. Then
\begin{equation}
  \tilde r_{i,t}
  =r_{i,t}+\frac{g_{\hat k_i}}{|W_i|}
    \mathds{1}[t\in W_i].
  \label{eq:compiled-credit}
\end{equation}
Thus $\sum_t(\tilde r_{i,t}-r_{i,t})=g_{\hat k_i}$: rank sets total credit and the box window conserves it while distributing it near estimated arrival. The auxiliary reward enters bank Bellman targets, while collection and evaluation retain the environment reward. Task-specific labels enter the shared compiler through $\rho_{\mathcal T}$.

The arrival midpoint also defines
\begin{equation}
  \omega_{i,t}=\exp\!\left[-\frac{(t-\bar t_i)^2}{2\sigma^2}\right],
  \qquad \sigma>0,
  \label{eq:arrival-weight}
\end{equation}
which provides a smooth arrival-proximity weight for sampling and behavior cloning. The box kernel fixes total credit; the Gaussian emphasizes actions near the midpoint.

Each accepted transition enters $\mathcal B_\Phi$ with training reward $\tilde r_{i,t}$, arrival weight $\omega_{i,t}$, and fixed sampling distribution
\begin{equation}
  p_\Phi(i,t)\propto
  1+\eta_r(\tilde r_{i,t}-r_{i,t})+\eta_a\omega_{i,t}.
  \label{eq:bank-priority}
\end{equation}
The distribution is normalized over all transitions in accepted records. Here $\eta_r,\eta_a\ge0$; the unit floor fixes the common priority scale. The resulting distribution retains episode context, emphasizes auxiliary credit and arrival proximity, and remains fixed throughout training.

\subsection{Online Learning with the Frozen Phase Bank}

\paragraph{Sampling and critic update}
Let $p_{\mathrm{on}}$ be uniform over $\mathcal B_{\mathrm{on}}$. A minibatch draws fraction $m\in[0,1]$ from the frozen bank, equivalently $p_{\mathrm{mix}}=m p_\Phi+(1-m)p_{\mathrm{on}}$. We use twin-critic SAC~\citep{haarnoja2018sac}, with $\tilde r$ on bank transitions and $r$ online.

\paragraph{Actor update}
Let $a_\theta^{\mathrm{eval}}(o)$ be the deterministic action used at evaluation and $Q_{\min,\psi}=\min_jQ_{\psi_j}$. Following demonstration-augmented RL~\citep{nair2018overcoming}, the Q-filter sets $\chi_{i,t}=1$ when bank action $a_t$ has no lower $Q_{\min,\psi}$ at $o_t$ than $a_\theta^{\mathrm{eval}}(o_t)$, and zero otherwise.
Let $\mathcal I_\Phi$ be the bank-origin samples in the current mixed minibatch and $\ell_{i,t}=d_a^{-1}\|a_\theta^{\mathrm{eval}}(o_t)-a_t\|_2^2$, the mean squared action deviation. Their weighted cloning loss is
\begin{equation}
  \mathcal L_{\mathrm{BC}}=
  \frac{\sum_{(i,t)\in\mathcal I_\Phi}
    \chi_{i,t}\omega_{i,t}\ell_{i,t}}
  {\sum_{(i,t)\in\mathcal I_\Phi}\chi_{i,t}\omega_{i,t}},
  \label{eq:bc-loss}
\end{equation}
We skip the BC term when the denominator is zero. Here $d_a$ is the action dimension. Because $\omega$ enters $p_\Phi$, arrival proximity controls both bank sampling and cloning strength.

Before online learning, the actor minimizes \Cref{eq:bc-loss} alone with $\chi\equiv1$. Online, let $\mathcal L_{\mathrm{SAC}}^{\mathrm{actor}}$ denote the standard entropy-regularized SAC actor loss evaluated on the mixed minibatch. We optimize
\begin{equation}
  \mathcal L_{\mathrm{actor}}
  =\mathcal L_{\mathrm{SAC}}^{\mathrm{actor}}
   +\lambda_{\mathrm{BC}}\mathcal L_{\mathrm{BC}},
  \label{eq:actor-loss}
\end{equation}
where $\lambda_{\mathrm{BC}}\ge0$ weights bank cloning. Frozen sampling, bank critic targets, and Q-filtered cloning integrate the phase record into SAC. \Cref{alg:phaserl} summarizes the procedure.

\endgroup

\begin{algorithm}[!t]
  \caption{Fixed phase-bank construction and online SAC}
  \label{alg:phaserl}
  \small
  \begin{algorithmic}[1]
    \Require $\mathcal D_{\mathrm{sup}}$, $(d_{\mathcal T},\mathcal Q_{\mathcal T},\rho_{\mathcal T})$, adapter $A_{\mathcal T}$, observer $f_\phi$, method settings
    \State Initialize $\mathcal B_\Phi$, $\mathcal B_{\mathrm{on}}$, and the SAC actor/critics
    \For{each support episode $\tau_i\in\mathcal D_{\mathrm{sup}}$}
      \State $(\hat q_i,\hat t_i,c_i)\gets f_\phi(A_{\mathcal T}(\tau_i),d_{\mathcal T},\mathcal Q_{\mathcal T})$ \Comment{one query}
      \State $\hat k_i\gets\rho_{\mathcal T}(\hat q_i)$
      \If{$c_i\ge c_{\min}$, $\hat k_i>0$, and $\hat t_i$ is valid}
        \State Compile $\tilde r$, $\omega$, and $p_\Phi$ by \Cref{eq:compiled-credit,eq:arrival-weight,eq:bank-priority}
        \State Insert $(o,a,\tilde r,o',b,\omega,p_\Phi)$ into $\mathcal B_\Phi$
      \EndIf
    \EndFor
    \State Freeze $\mathcal B_\Phi$; warm-start by weighted BC with $\chi\equiv1$
    \For{each online iteration}
      \State Collect with $\pi_\theta$ into $\mathcal B_{\mathrm{on}}$
      \State Draw a mixed minibatch with bank fraction $m$ and distribution $p_\Phi$
      \State Update standard SAC critics using $\tilde r$ on bank and $r$ online
      \State Update the actor by \Cref{eq:actor-loss}
    \EndFor
  \end{algorithmic}
\end{algorithm}

% =================================================================
\section{SurgPhaseBench}
\label{sec:platform}
% =================================================================

SurgPhaseBench makes partial progress an explicit evaluation object for sparse-reward surgical manipulation. Each environment couples an executable terminal objective with task-authored visual milestones, so an episode can be evaluated not only by whether it succeeds but also by which milestone it reaches and when that evidence first appears. Built on ManiSkill3~\citep{tao2025maniskill3}, its twelve environments span rigid and Warp-based deformable interaction~\citep{macklin2022warp}.

\begin{figure*}[!t]
  \centering
  \includegraphics[width=\textwidth]{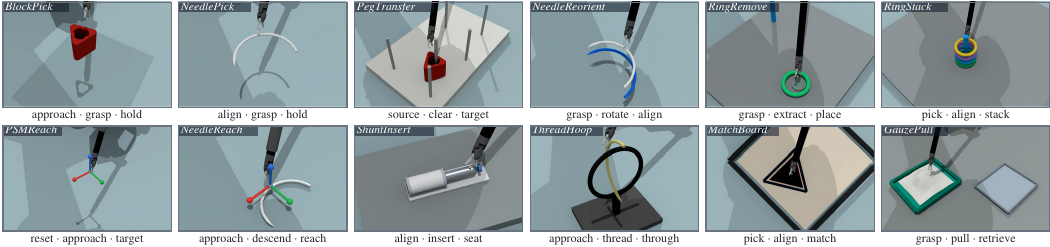}
  \caption{One phase-structured contract spans twelve executable surgical environments. Five contact-rich tasks test the complete observer--compiler--learner loop; seven additional tasks establish execution and interface breadth.}
  \label{fig:platform-panel}
\end{figure*}

\subsection{Phase-Structured Task Interface}

The benchmark separates task semantics from evaluation structure. Each task supplies a short description, an ordered phase vocabulary, and one-line visual decision rules. Task-specific labels preserve distinctions such as grasp--lift--transfer, align--seat, and attach--pull, while their ordinal mapping lets the same observer and compiler operate across tasks. All environments share the same patient-side arm model, delta end-effector position control, state observations, and terminal sparse success. The five policy tasks use horizons of 75, 75, 90, 90, and 220 steps in task order. BlockPick, NeedlePick, NeedleReach, and ShuntInsert include the same proximity grasp assist for every compared method; GauzePull instead uses a Warp soft-body attachment model.

\subsection{Task Suite and Benchmark Scope}

The twelve tasks span reaching, grasp-and-lift, transfer, insertion, threading, reorientation, stacking, and deformable pulling (\Cref{fig:platform-panel}). BlockPick, NeedlePick, PegTransfer, ShuntInsert, and GauzePull form the contact-rich policy-learning track: four cover rigid-object interaction, while GauzePull places deformable attachment and pulling under the same phase contract. The remaining seven environments establish executable and semantic breadth rather than entering the policy aggregate. Accordingly, the task atlas documents suite coverage, the five-task track supplies policy-learning evidence, and physical studies test external validity separately.

% =================================================================
\section{Experiments}
\label{sec:experiments}
% =================================================================

We test whether the visual interface recovers phase and arrival, whether its records improve finite-budget policy learning, and how support quality and temporal localization affect that gain. Simulation controls are followed by observer transfer and closed-loop learning on hardware.

\subsection{Simulation Setup}
\label{sec:simulation-benchmark}
\label{sec:bringup}

\paragraph{Evaluation protocol}
All simulation studies run on the five-task policy-learning track. Support is collected before the downstream learner and then frozen. We study rollouts from a checkpoint trained with privileged dense shaping (\emph{shaped}), a checkpoint trained only with terminal sparse reward (\emph{sparse}), and untrained random exploration (\emph{random}). Privileged shaping is therefore confined to acquisition of the shaped support pool; the observer sees only its storyboards, and the downstream learner and evaluation retain the common observation/action interface and sparse environment reward. Within each comparison, methods receive identical support rollouts, learner settings, and evaluation budgets.

The evidence is organized into three independently trained protocols: the primary feedback benchmark, 120k-update diagnostics for compiler components and support source, and two 500k-transition matched controls. The primary and diagnostic studies share the 120k SAC configuration, but their absolute values are compared only within the corresponding figure or table. Because horizons and collection parallelism differ, all methods in the primary comparison are read at pre-specified task-local points---250k, 375k, 275k, 300k, and 175k environment transitions, respectively---and averaged without weighting. Every condition uses eight seeds; task values are seed means and curves show 95\% CIs.

\paragraph{Baselines and controls}
We compare three learned-feedback families that use only image observations and a task description. MultiTile-CLIP applies CLIP-RM~\citep{radford2021clip,rocamonde2024zeroshot} to our seven-tile views; BT-Pref~\citep{christiano2017preferences,wang2024rlvlmf} compares pairs; and Likert-RM~\citep{luu2025erlvlm} rates an attempt with a scalar. We also report no-feedback SAC, single-view CLIP-RM, and privileged Oracle-Support with ground-truth phase labels for the same pool.

\Cref{tab:matched-controls-main} reports two independent 500k-transition controls. In panel (a), GPT-5.4 Scalar and Phase+Arrival share the model, seven-image input, shaped support, learner, evaluation, and annotation budget (960 requests; 134k visual tokens); only the output schema changes. Panel (b) gives MultiTile-CLIP the same phase-filtered BC, $\pm10$-step credit window, and phase-aware SAC integration used by Phase+Arrival, testing whether a matched learning stack closes the gap. This second study applies the same task-local BC/gate refinement to both ShuntInsert rows. Compare only within panels.

\paragraph{Implementation details}
\label{sec:implementation}
The fixed adapter forms seven panels per episode: six chronological composites on an even temporal grid and one start/final motion-difference panel. Each timed composite contains the full view, fixed task-specified crops, and its environment-step label. The observer is GPT-5.4 at temperature zero with a 400-token cap and no task-specific fine-tuning. All tasks use gate $c_{\min}{=}0.70$, total-credit schedule $g=(0,0,0.20,0.55,0.55,0.90)$, Gaussian width $\sigma{=}8$, a $\pm10$-step arrival window, and priority coefficients $(\eta_r,\eta_a)=(4,1)$.

The actor is warm-started for 5k updates; online learning uses bank mix $m{=}0.6$ and BC weight $\lambda_{\mathrm{BC}}{=}0.6$. SAC uses entropy autotuning, $\gamma{=}0.80$ ($0.96$ for GauzePull), a 200k replay buffer, batch size 256, AdamW at $3\times10^{-4}$, and target rate $0.01$. Each seed uses an RTX~5080 Laptop GPU and 8--32 parallel environments, set by task horizon and cost.

\begin{figure}[!b]
  \centering
  \includegraphics[width=0.99\columnwidth]{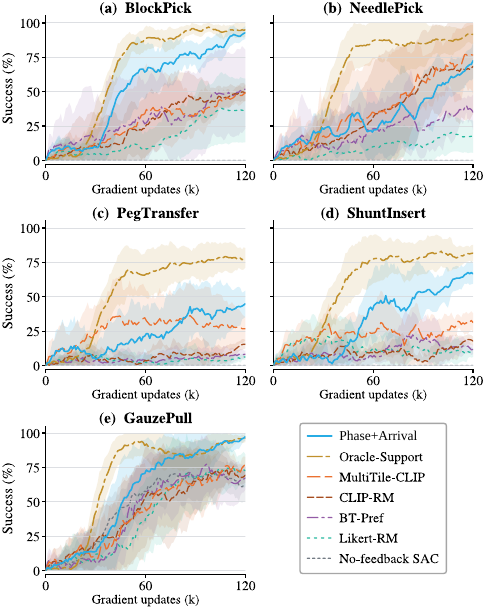}
  \caption{Primary five-task feedback comparison after 120k SAC updates (eight seeds; shading, 95\% CIs). At synchronized task-specific reporting points, Phase+Arrival averages 75.2\% versus 52.1\% for MultiTile-CLIP using the same views and leads on four tasks. Sparse reward reaches 12.3\%; Oracle-Support reaches 88.6\% using ground-truth phases on the same identically compiled pool, providing the pool's performance ceiling.}
  \label{fig:core-learning-curves}
\end{figure}

\begin{figure*}[!t]
  \centering
  \includegraphics[width=\textwidth]{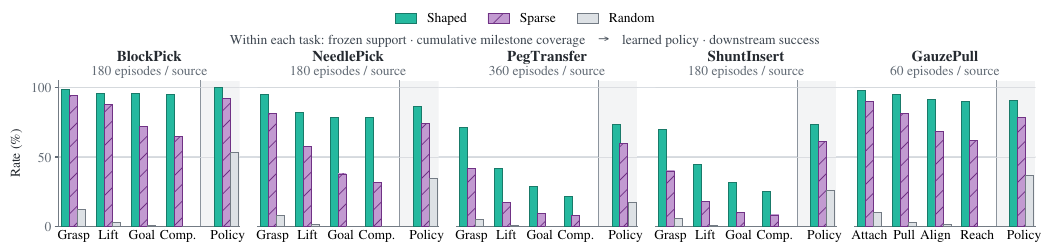}
  \caption{Independent support-source diagnostic. For each task, the first four groups report cumulative milestone coverage in shaped, sparse, and random support, while the separated Policy group reports success after 120k updates (eight seeds per condition). Each source contains 960 episodes across tasks, giving 2,880 observer queries and 120 policy runs. PegTransfer exposes the partial-progress regime: only 21.7\% of shaped episodes complete, yet the learned policy reaches 73.4\% success.}
  \label{fig:pool-source}
\end{figure*}

\subsection{Simulation Results}
\label{sec:simulation-results}

\Cref{fig:core-learning-curves} compares seven feedback conditions on the five-task track. At the synchronized reporting points, Phase+Arrival averages $75.2\%$ success against $52.1\%$ for matched-view MultiTile-CLIP, $35.0\%$ for BT-Pref, and $27.9\%$ for Likert-RM, leading on four tasks. NeedlePick is the exception ($72.5\%$ versus $75.6\%$), although the phase-conditioned method later attains the highest learned-feedback peak on every task; we report synchronized points throughout. The same figure adds the wider family: sparse reward alone reaches $12.3\%$, standard CLIP-RM $44.0\%$, and privileged Oracle-Support $88.6\%$, leaving a $13.4$-point support-observation gap.

\begin{table}[!b]
  \caption{Two independent 500k-transition controls (eight-seed mean success, in \%). (a) GPT-5.4 and input are fixed; only scalar versus phase+arrival (P+A) output changes. (b) MultiTile-CLIP (MT-CLIP) receives the matched learning stack; both rows use the same ShuntInsert refinement. $\Delta$ is within-panel.}
  \label{tab:matched-controls-main}
  \centering
  \scriptsize
  \setlength{\tabcolsep}{1.25pt}
  \renewcommand{\arraystretch}{1.45}
  \begin{tabular*}{\columnwidth}{@{\extracolsep{\fill}}lrrr@{\hspace{2pt}}rrr@{}}
    \toprule
    & \multicolumn{3}{c}{\textit{(a) Output format isolated}} &
      \multicolumn{3}{c}{\textit{(b) Learning stack matched}} \\
    \cmidrule(lr){2-4}\cmidrule(lr){5-7}
    Task & GPT sc. & P+A & $\Delta$ & MT-CLIP & P+A & $\Delta$ \\
    \midrule
    BlockPick   & 90.0 & \textbf{93.8} & +3.8  & 88.0 & \textbf{93.8} & +5.8 \\
    NeedlePick  & 82.0 & \textbf{85.4} & +3.4  & 78.0 & \textbf{85.4} & +7.4 \\
    PegTransfer & 68.0 & \textbf{73.4} & +5.4  & 66.0 & \textbf{73.4} & +7.4 \\
    ShuntInsert & 58.0 & \textbf{78.0} & +20.0 & 61.0 & \textbf{81.0} & +20.0 \\
    GauzePull   & 86.0 & \textbf{90.6} & +4.6  & 87.0 & \textbf{90.6} & +3.6 \\
    \midrule
    \textbf{Average} & \textbf{76.8} & \textbf{84.2} & \textbf{+7.4} &
      \textbf{76.0} & \textbf{84.8} & \textbf{+8.8} \\
    \bottomrule
  \end{tabular*}
\end{table}

In \Cref{tab:matched-controls-main}(a), changing only GPT-5.4's output from a scalar to phase and first arrival raises average success from $76.8\%$ to $84.2\%$. In panel (b), Phase+Arrival remains ahead after MultiTile-CLIP receives the matched learning stack ($84.8\%$ versus $76.0\%$); the panels are independent.

For the five-task shaped pool, 960 seven-image requests cost about \$7.2. Using the per-task episode counts in \Cref{fig:pool-source} and the stated horizons, full-horizon scoring would process 88{,}800 images instead of 6,720 ($13.2\times$ more).

\paragraph{Observer reliability}
We compare observer outputs with frozen simulator-derived maximal-phase and first-arrival references. \emph{Valid} denotes a parseable structured record, while \emph{gated} denotes a valid record retained by the confidence, progress, and timing gate in \Cref{sec:hindsight-integration}. Across 960 episodes, valid/gated rates are $98.6$--$100.0\%$/$87.8$--$96.7\%$ and within-one agreement is $73.3$--$91.1\%$; exact agreement is $33.3$--$58.3\%$. BlockPick is weakest within one phase, while ShuntInsert is hardest exactly because seating is largely hidden. First-arrival MAE against the reference step is 8.4--24.8 steps; the compiler distributes credit within $\pm10$ steps of the predicted arrival rather than requiring an exact event frame.

\begin{table}[!b]
  \caption{NeedlePick mechanism study at 120k updates (eight-seed mean success, in \%). (a) Cumulative construction; (b) independently trained alternatives. All rows share support pools and evaluation; the full compiler is repeated as a reference.}
  \label{tab:needlepick-mechanism}
  \centering
  \scriptsize
  \setlength{\tabcolsep}{1.0pt}
  \renewcommand{\arraystretch}{1.00}
  \begin{tabular*}{\columnwidth}{@{\extracolsep{\fill}}lrr@{\hspace{2pt}}lrr@{}}
    \toprule
    \multicolumn{3}{c}{\textit{(a) Cumulative build-up}} &
      \multicolumn{3}{c}{\textit{(b) Alternative mechanisms}} \\
    \cmidrule(lr){1-3}\cmidrule(lr){4-6}
    Method & Shaped & Sparse & Method & Shaped & Sparse \\
    \midrule
    Uniform-pool BC   & 30.2 & 26.4 & Random SAC          & 5.0  & 5.0  \\
    Phase-weighted BC & 50.4 & 42.0 & Reward only         & 12.6 & 9.8  \\
    Online SAC + BC   & 62.8 & 54.4 & Direct phase reward & 56.2 & 51.8 \\
                      &      &      & HER                 & 56.8 & 52.6 \\
    \textbf{Full compiler} & \textbf{86.8} & \textbf{74.0} &
      \textbf{Full compiler} (ref.) & \textbf{86.8} & \textbf{74.0} \\
    \bottomrule
  \end{tabular*}
\end{table}

% Queue the full-width physical-evidence figures early enough to place them at
% the top of the physical-experiment page rather than leaving a float gap.
\begin{figure*}[!t]
  \centering
  \includegraphics[width=\textwidth]{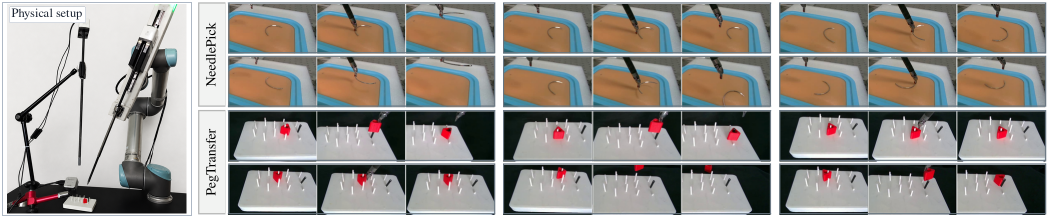}
  \caption{Physical experimental setup and observer examples. Left: a UR5 carries a da Vinci Xi EndoWrist instrument through a custom adapter and is viewed by a fixed oblique RealSense D405. Right: success, partial, and failure examples from teleoperated physical needle-picking and peg-transfer videos; observer agreement across all three physical tasks is reported in the text.}
  \label{fig:physical-observer}
\end{figure*}

\begin{figure*}[!t]
  \centering
  \includegraphics[width=\textwidth]{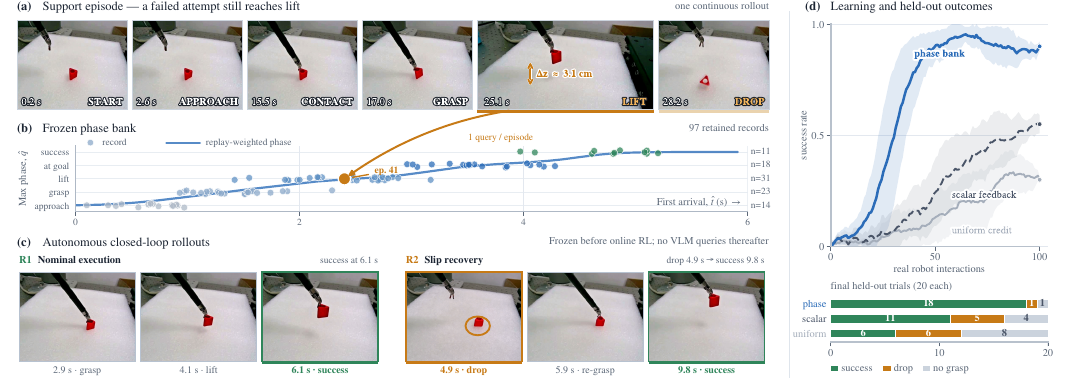}
  \caption{Physical robot block-picking closed-loop experiment. (a) A failed support episode reaches lift before dropping the block. (b) One offline query per episode produces a 97-record phase bank, frozen before online learning. (c) Representative autonomous rollouts, including slip recovery. (d) After 100 online transitions, phase-conditioned replay achieves 18/20 held-out successes, versus 11/20 for framewise scalar feedback and 6/20 for uniform credit; deployment uses no VLM queries.}
  \label{fig:realblockpick-closed-loop}
\end{figure*}

\subsection{Ablation Studies}
\label{sec:mechanism}

\label{sec:support-coverage}
Each source uses the same 960-episode allocation (180/180/360/180/60 in task order) and 960 queries. Shaped and sparse collectors each use about 500k aggregate transitions per task over eight seeds; random support uses no trained collector. In this independent 120k-update diagnostic, the pools average $84.8\%$, $73.3\%$, and $33.7\%$ success (\Cref{fig:pool-source}); these are not the primary results in \Cref{fig:core-learning-curves}. PegTransfer illustrates the partial-progress regime: only $21.7\%$ of shaped rollouts finish, while $71.7\%/41.7\%/29.2\%$ reach grasp/lift/at-goal and support $73.4\%$ downstream success. Random support has no successes and $\leq1.7\%$ at-goal coverage.

ShuntInsert provides a complementary case: only $25.0\%$ of shaped episodes complete, while $70.0\%/45.0\%/31.7\%$ reach grasp/lift/at-goal. Its $20.0$-point improvement in both matched-control panels is also the largest in \Cref{tab:matched-controls-main}. Together, these tasks show that terminal collector success is not a sufficient support statistic: a pool may be weak at completion yet rich in transferable early- and mid-stage behavior.

\Cref{tab:needlepick-mechanism} separates behavior selection from temporal credit. Uniform-pool BC clones all support; Phase-weighted BC gates and weights it; Online SAC+BC adds sparse-reward SAC; and Full compiler additionally supplies record-conditioned credit, weights, and priorities. Phase weighting contributes $20.2/15.6$ points on shaped/sparse support, and full compilation adds $24.0/19.6$ over Online SAC+BC. Reward only starts without BC, Direct phase reward omits localization and record-derived weights, and HER relabels target positions after Uniform-pool BC without a phase bank. Direct phase reward loses $30.6/22.2$ points; HER is similar, but only phase records expose event time.

Together, source and mechanism studies expose complementary limits. Coverage determines which milestones enter the bank; compilation selects and credits them at first arrival. Phase magnitude alone cannot recover the full gain, and localization cannot recover a phase absent from support.

\subsection{Physical Robot Experiments}
\label{sec:physical-validation}

Physical experiments use a custom patient-side manipulator: a 6-DoF UR5 carries a da Vinci Xi EndoWrist Mega Needle Driver through a 5-DoF adapter (\Cref{fig:physical-observer}). A fixed oblique RealSense D405 provides the online RGB-D target pose and synchronized diagnostic video ($848\times480$, $\approx22$\,fps). The observer returns per-phase arrival times in seconds; the maximal phase and its timestamp form a degenerate bracket for the same downstream record used in simulation.

We test observer transfer from simulation to teleoperated physical videos of block picking, needle picking, and peg transfer. Block picking evaluates autonomous closed-loop learning in a 20-trial study. A human labels maximal phase and outcome before the VLM scores 15 videos per task (five success, five partial, five failure). All 45 outputs are valid; exact and within-one agreement reach $71.1\%$ and $91.1\%$, and 41/45 success labels match human labels. Peg transfer is hardest because its phases look similar from one view.

In the physical block-picking study, the policy maps target pose and joint state to $a_t=(\Delta x,\Delta y,\Delta z,g)$ (\Cref{fig:realblockpick-closed-loop}). One query per episode yields a frozen bank of 97 records. Each condition receives the same support and 100 online transitions, followed by 20 held-out trials. Framewise feedback scores each frame without phase or arrival; uniform credit preserves rank but spreads credit and replay weight across the episode.

Phase replay succeeds in 18/20 trials, versus 11/20 for scalar feedback and 6/20 for uniform credit. A retained failure lifts $3.1$\,cm, and an online rollout recovers from a slip. Under the same 100-transition budget, phase replay leaves one drop and one no-grasp, versus five/four for scalar feedback and six/eight for uniform credit. The block-picking study provides end-to-end evidence; the needle-picking and peg-transfer video audits test observer transfer across tasks.

% Keep all physical-experiment evidence visually ahead of the discussion.
\FloatBarrier

% =================================================================
\section{Discussion}
\label{sec:discussion}
% =================================================================

The matched controls and ablations clarify what the structured record contributes. Holding the VLM, storyboard, support, and learning stack fixed, phase-and-arrival feedback remains more effective than a scalar summary, so the gain is not explained by additional visual evidence or an unmatched learner. The support-source and mechanism studies then separate two roles: milestone coverage determines which partial behaviors are available for reuse, while first arrival specifies the transition neighborhood where their credit should concentrate. A more precise observer cannot supply behavior missing from support, and broad support alone does not determine where its credit should land. This separation localizes failures: missing phases call for new support, whereas uncertain timing calls for calibration.

Freezing records before learning makes the interface inspectable: each episode can be traced to its phase, confidence, and arrival bracket, and the same bank can be recompiled without querying the VLM during deployment. The physical studies separate observer transfer across tasks from end-to-end closed-loop learning in physical block picking. More broadly, the approach is most natural for procedures with ordered, visually distinguishable milestones. Support-aware acquisition and calibrated uncertainty offer routes to ambiguous phases, longer horizons, and deformable physical tasks.

% =================================================================
\section{Conclusion and Future Work}
\label{sec:conclusion}
% =================================================================

This work presents phase-and-first-arrival feedback as a reusable interface between trajectory-level visual supervision and sparse-reward control. Phase identifies useful partial attempts, while first arrival localizes their replay, critic credit, and behavior reuse. Across simulation and hardware, the results show that a frozen structured record can preserve partial progress and support closed-loop learning without deployment-time VLM queries. Future work will extend support-aware acquisition and uncertainty calibration to raw-image policies, deformable interaction, and longer tasks.

% =================================================================

\bibliographystyle{IEEEtran}
\balance
\bibliography{main}
\end{document}